%% file: main.tex
\documentclass[10pt,twocolumn,letterpaper]{article}

\usepackage{cvpr}              

\input{preamble}

\definecolor{cvprblue}{rgb}{0.21,0.49,0.74}
\usepackage[pagebackref,breaklinks,colorlinks,citecolor=cvprblue]{hyperref}
\usepackage{cite}
\usepackage[switch]{lineno}
\usepackage{booktabs}
\usepackage{tabularray}
\usepackage{caption}
\usepackage{subcaption} 
\usepackage{newfloat}
\usepackage{colortbl}
\usepackage{multirow}
\usepackage{url}
\usepackage{verbatim}
\usepackage{graphicx}
\usepackage{color}

\usepackage{makecell}
\usepackage[ruled,vlined]{algorithm2e}

\usepackage{hyperref}
\hypersetup{
    colorlinks=true,
    linkcolor=blue,
    filecolor=magenta,      
    urlcolor=blue,
    pdftitle={Overleaf Example},
    pdfpagemode=FullScreen,
    }

\definecolor{commentcolor}{RGB}{110,154,155}   

\def\confName{CVPR}
\def\confYear{2024}

\title{Learning to Rank Patches for Unbiased Image Redundancy Reduction}

\author{
\textbf{Yang Luo$^1$, Zhineng Chen$^{1 \thanks{Corresponding author: Zhineng Chen}}$ , Peng Zhou$^2$ , Zuxuan Wu$^1$ , Xieping Gao$^3$ , Yu-Gang Jiang$^1$ } \\
\normalsize{$^1$School of Computer Science, Shanghai Collaborative Innovation Center of Intelligent Visual Computing, Fudan University}  \\ 
\normalsize{$^2$University of Maryland, College Park } \\
\normalsize{ $^3$College of Information Science and Engineering, Hunan Normal University}  \\
{\tt\small yangluo21@m.fudan.edu.cn, \{zhinchen, zxwu, ygj\}@fudan.edu.cn,} \\ {\tt\small pengzhou@terpmail.umd.edu, xpgao@hunnu.edu.cn}
}

\begin{document}
\maketitle
\begin{abstract}
Images suffer from heavy spatial redundancy because pixels in neighboring regions are spatially correlated. Existing approaches strive to overcome this limitation by reducing less meaningful image regions. However, current leading methods rely on supervisory signals. They may compel models to preserve content that aligns with labeled categories and discard content belonging to unlabeled categories. This categorical inductive bias makes these methods less effective in real-world scenarios. To address this issue, we propose a self-supervised framework for image redundancy reduction called \textbf{L}earning \textbf{t}o \textbf{R}ank \textbf{P}atches (LTRP). We observe that image reconstruction of masked image modeling models is sensitive to the removal of visible patches when the masking ratio is high (e.g., 90\%). Building upon it, we implement LTRP via two steps: inferring the semantic density score of each patch by quantifying variation between reconstructions with and without this patch, and learning to rank the patches with the pseudo score. The entire process is self-supervised, thus getting out of the dilemma of categorical inductive bias. We design extensive experiments on different datasets and tasks. The results demonstrate that LTRP outperforms both supervised and other self-supervised methods due to the fair assessment of image content. Code is available at \url{https://github.com/irsLu/ltrp}.
\end{abstract}

\section{Introduction}
\begin{figure}[!htp]
   \centering
    \includegraphics[width=1\columnwidth]{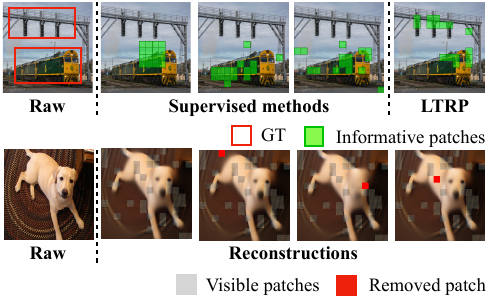}
     \caption{\textbf{Redundancy reduction} (upper): given an image, LTRP selects informative patches regardless of whether they are from the categories already learned. While for supervised methods (from left to right: GFNet \cite{wang2020glance}, Grad-CAM \cite{selvaraju2017grad} using ViT, EViT \cite{liang2022not}) the preserved patches are mostly located on the learned category. \textbf{Patch removing} (bottom): an image and its reconstructions using MAE \cite{he2022masked} with different sets of visible patches, columns 3-5 denote that the \emph{red} patch is removed from the visible set. They generate reconstructions with different levels of semantic shift, e.g., without the dog tail, eyes, etc. }
    \label{fig:intuition}
\end{figure}

Image redundancy reduction refers to discarding the less informative regions of an image while preserving its essential semantics. It is a fundamental task in computer vision as compact image representation is an attractive property in various applications, especially in the era of vision Transformer (ViT) \cite{DBLP:conf/iclr/DosovitskiyB0WZ21,liu2021swin}. The computational complexity of ViT grows quadratically with the number of image patches, making patch-level image redundancy reduction gaining increasing attention.

Current leading methods \cite{wang2020glance,cordonnier2021differentiable,yang2020resolution,pan2021ia} generally require supervised learning to endow the model with the capability to recognize informative image patches. Specifically, the model is simultaneously trained to estimate the patch's informativeness and classify the image to the right category. Thus, the learning process tends to select image patches that are less likely to degrade the classification accuracy, leading to categorical inductive bias. The bias may cause the model to remove image patches containing meaningful semantics that do not belong to the learned category. As shown in the upper of \cref{fig:intuition}, supervised methods trained based on ImageNet-1K tend to select informative patches located on \emph{train}, a category learned. While considering patches on unseen categories like \emph{traffic light} as redundant and removing them. Note that such categorical inductive bias is unavoidable for models trained based on famous image datasets \cite{deng2009imagenet,lin2015microsoft,Everingham15}, which typically annotate one or a few categories for each image. On the other hand, activation mapping of popular unsupervised pre-training methods \cite{peng2022crafting,caron2021emerging} also reveal the informativeness of image patches to some extent. Nevertheless, they are less effective as the pre-training is not targeted for redundancy reduction. 

Recently, masked autoencoder (MAE) \cite{he2022masked} has gained increasing attention due to its great representative learning capability. Given an image with a portion of patches randomly masked, MAE learns the feature representation by reconstructing the masked patches. It can generate a reconstruction with meaningful semantics by using a small subset (e.g., 25\%) of visible patches. We observe that when the masking ratio is high (e.g., 90\%), further removing a visible patch may cause an obvious semantic shift in the reconstructed image. As shown in the bottom of \cref{fig:intuition}, if the only visible patch on the dog's tail is masked, the reconstruction will not have the tail. Conversely, masking a patch on the dog's body does not induce perceptible semantic alteration. The result suggests that the informativeness of patches can be revealed from such an unsupervised perspective. 


Motivated by this observation, we propose learning to rank patches (LTRP), a self-supervised framework to fairly reduce image redundancy. It firstly leverages a pre-trained MAE to generate a pseudo score for each visible patch by quantifying the semantic difference between reconstructions with and without that patch. 
A larger pseudo score indicates that the patch has a greater impact on the quality of the reconstructed image and is thus more valuable. Therefore, LTRP utilizes another ranking model to learn to rank the visible patches according to the pseudo scores, which successfully learns this scoring property. After pre-training, by ranking the patches descendingly according to their scores and keeping the top ones, LTRP can remain valuable patches for the subsequent downstream tasks like classification. As shown in the top right image of \cref{fig:intuition}, patches containing meaningful semantics are unbiasedly preserved under the same keep ratio.


We conduct extensive experiments to validate the unbiasedness of LTRP in image redundancy reduction. For image classification at different keep ratios, LTRP not only achieves competitive accuracy compared to current leading methods in single-label classification, but also outperforms them by a clear margin in multi-label classification. When inspecting categories not included in ImageNet-1K, patches remained by LTRP are also more consistent with object detection and semantic segmentation labels, getting superior performance in all the evaluated metrics. Moreover, if a small ranking model is employed, we observe that an approximate linear speedup with the keep ratio is coincidentally obtained for image classification, which constitutes a promising solution for efficient ViT. Contributions of this paper are summarized as follows:

\begin{itemize}
  \item We propose to treat the image redundancy reduction as a self-supervised patch-level ranking problem. It eliminates the reliance on supervisory signals, expanding the means to reduce image redundancy. 
  \item We develop LTRP. It quantifies the variation between MAE-reconstructed images to create pseudo labels, then learns to rank these patches. It offers an elegant way to avoid categorical inductive bias.
  \item Extensive quantitative and qualitative evaluations demonstrate that LTRP can fairly and effectively score the learned and unseen categories, and provide highly competitive solutions for efficient ViT.
\end{itemize}

\begin{figure*}[htb]
\centering
\includegraphics[width=\linewidth]{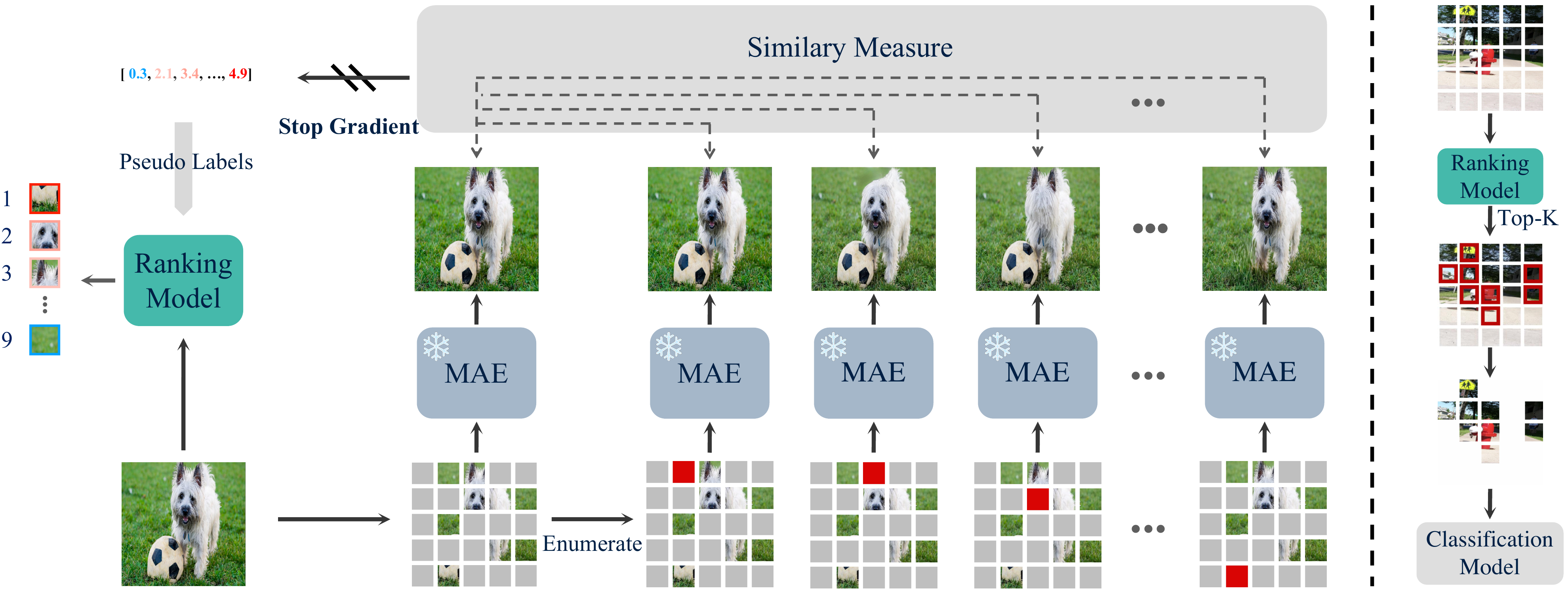}
\caption{\textbf{LTRP training (left):} given an image, LTRP randomly selects a set of visible patches using a high masking ratio (e.g. 90\%). A pre-trained MAE (parameter frozen) is applied to get its reconstruction, i.e., the anchor image. Then, the visible patches (red) are removed one by one with replacement, each time generating a new reconstruction and its semantic density score w.r.t the anchor image. The scores are treated as pseudo labels to train the ranking model using learning to rank. Once trained, the ranking model is preserved and the MAE is discarded. \textbf{LTRP inference (right):} Top-k patches are selected using the trained ranking model and fed into downstream tasks.}
\label{fig:idea}
\end{figure*}

\section{Related Work}
\subsection{Image Redundancy Reduction}
\label{sec:image_red2}
While sharing similarities with tasks like salient object detection \cite{duan2011visual,jiang2013saliency} and image compression \cite{rabbani1991digital}, image redundancy reduction has been studied mainly in the era of deep learning. Some CNN-based methods explored this task for inference efficiency \cite{wang2020glance,cordonnier2021differentiable,yang2020resolution}. Meanwhile, ViT models have been implicitly endowed with the redundancy reduction capability. Based on how this capability is built, we can broadly categorize existing studies into three classes. 


\textbf{Class activation mapping (CAM):} 
Starting from a label, CAM \cite{zhou2016learning,selvaraju2017grad} was proposed to provide visual explanations for CNNs by combining feature maps from the penultimate layer using learned or gradient-based weights. CAM achieved success in various visual tasks, such as weakly-supervised object detection \cite{9069411} and semantic segmentation \cite{ahn2018learning,fan2020learning}, which regards the activation as initial pseudo object labels. CAM can also be applied to ViT. However, it relies on supervisory signals to build the localization ability, thus suffering from the categorical inductive bias.

\textbf{Self-supervised activation mapping:} 
Recent studies in self-supervised learning (SSL) shows that the capability to perceive fine-grained objects can be obtained from contrastive learning-based pre-training. For example, object locations can be roughly inferred by using MoCo \cite{chen2020improved} to train a CNN and visualizing activation mapping of the last convolutional layer \cite{peng2022crafting}. A similar capability is also observed from the heads of DINO \cite{caron2021emerging}. Object localization is a natural property learned from SSL. However, as redundancy reduction is not the central goal for SSL, preserving image patches according to the generated activation mapping generally leads to worse performance in downstream tasks.


\textbf{Token reduction:}
Since the computational complexity of ViT is quadratically linear to the number of tokens \cite{DBLP:conf/iclr/DosovitskiyB0WZ21}, reducing tokens, which can be viewed as a special form of image patches by using token-to-patch mapping, is an effective method for efficient ViT. Current studies can be classified into two categories. First, token pruning that discards redundant tokens. It involves retraining a network to dynamically filter tokens \cite{rao2021dynamicvit}, using the policy gradient to select patches \cite{pan2021ia}, reducing the number of tokens as inference proceeds \cite{yin2022vit}, etc. Second, token merging that combines similar tokens. It involves leveraging attention scores between the class token and other tokens to gradually merge tokens \cite{liang2022not}, merging tokens between two subsets with equal size by using similarity of their key vectors from the self-attention layer \cite{bolya2022token}  and others \cite{xu2022groupvit,renggli2022learning}. Typically, the research focus of these methods is on efficient ViT. Unbiased redundancy reduction is rarely considered. 



\subsection{Learning to Rank}
Learning to rank (LTR) \cite{liu2009learning} is a technique widely employed in information retrieval tasks such as search \cite{chapelle2011yahoo} and recommendation \cite{karatzoglou2013learning}. LTR primarily involves three types according to the utilization of document labels: the point-wise method \cite{caruana1995using,li2007mcrank}, which uses the rank of an individual document as label and formulates the learning as either classification or regression problem; the pair-wise method \cite{cao2006adapting,burges2005learning}, which utilizes the relative order of two documents as label for learning; and the list-wise method \cite{cao2007learning,xia2008listwise}, which considers the order relationship among all or a part of documents associated with a given query. Some studies \cite{xu2016pairwise,liu2017rankiqa} have made meaningful attempts by leveraging learning to rank in computer vision tasks. To our knowledge, it has not been used for image redundancy reduction before.

LTRP is an application of LTR to image redundancy reduction. The query is the randomly masked image. Each document is associated with a visible patch, and whose pseudo score is the document label. Different from most existing solutions where the label is obtained from manual annotation or human-machine interaction, our label is obtained from an elegantly designed pre-task, thus enabling a novel self-supervised solution as described below. 


\section{Method}

An overview of our learning to rank patches (LTRP) is depicted in \cref{fig:idea}. Its core elements can be summarized using two words: infer and rank. Given a pre-trained MAE model and an image with a portion of patches randomly masked, LTRP first \emph{infers} the semantic density score for each visible patch by quantifying variation between the reconstructions with and without that patch. By repeating this process patch-by-patch, we obtain a series of scores. Then, LTRP \emph{ranks} these patches by treating their scores as pseudo labels using learning-to-rank algorithms. When the training converges, we use the ranking model to select informative patches and only consider them in downstream tasks. 



\subsection{Preliminary}
MAE \cite{he2022masked} learns a robust image representation by first masking a portion of image patches and then recovering the original image, namely the mask-then-prediction task. Specifically, MAE divides an image \(I\) into visible and masked patch sets according to a predefined masking ratio, where the visible set \(\mathbf{P}_v=\left[\mathbf{p}_1, \mathbf{p}_2, \ldots, \mathbf{p}_n\right]\) has \(n\) patches. Then, the pre-training target can be written as: 

\begin{equation}
    \underset{\theta}{min} H(\mathcal{T}(I_{m}), I_{m}^R),I_{m}^R= \mathcal{M}_{\theta}(\mathbf{P}_v)
\end{equation}

\noindent where \(I_{m}\), \(I_{m}^{R}\) and \(\mathcal{M}_{\theta}\) denote the masked image, the reconstructed image and MAE parameterized by \(\theta\), respectively, \(\mathcal{T}\) denotes a certain transformation (optional, e.g., HOG \cite{Wei_2022_CVPR}, normalized pixel values of each masked patch \cite{he2022masked}, etc.), and \({H}\) is a distance measurement function.


\subsection{Patch Informativeness Inference}

To infer the patch informativeness, we adopt a higher masking ratio of 90\% rather than 50\% or 75\% used in previous MAE studies \cite{he2022masked,Xie_2022_CVPR}. The reason is twofold. First, it enables reconstructions exhibiting perceptible change even when only one visible patch is further removed. Second, it decreases the likelihood of semantically similar patches being visible simultaneously, which may confuse the model regarding their informativeness. 

Then, we use a pre-trained, parameter-frozen MAE to perform the reconstruction. For an image, the visible set \(\mathbf{P}_v\) is stochastically selected and its reconstruction \(I_{m}^{R}\), or saying, the anchor image, is obtained, we iteratively remove a visible patch \(\mathbf{p}_i \in \mathbf{P}_v\) to obtain \(\mathbf{P}_{v}^{i}=\mathbf{P}_v\backslash\mathbf{p}_i\). It can be simply achieved by using masked attention \cite{vaswani2017attention}. By reconstructing again, we get \(I_{m}^{R_{i}}\), the image reconstruction without the \(i\)-th visible patch. A similarity-based function \(\mathcal{F}\) runs over the pair \(<I_{m}^{R}, I_{m}^{R_{i}}>\), and the quantitative similarity \(y_i\) is calculated by:
 \begin{equation}
y_i=\mathcal{F}(I_{m}^{R}, I_{m}^{R_{i}}), I_{m}^{R_{i}}=\mathcal{M}_{\theta}(\mathbf{P}_{v}^{i})
\label{eq2}
\end{equation}

We term the inferred result as the semantic density score, as it quantizes the semantic variant between the two reconstructions. We simply use the \(\ell_1\) distance to measure the variant in this paper. By polling all the visible patches one-by-one, we get a pseudo score vector \(\mathbf{y}=\left[y_1, y_2, ...,y_n\right]\). Note that the scores are obtained in a self-supervised manner, and \cref{eq2} can be regarded as the pre-task to reveal the patch informativeness. In the implementation, the score inference can be parallelized for training acceleration.


\begin{figure}[!htp]
\centering
\includegraphics[width=\linewidth]{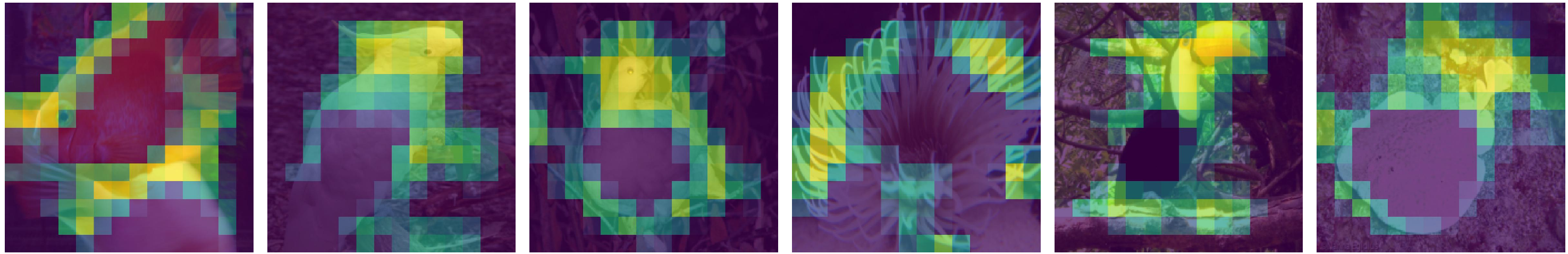}
\caption{Normalized score maps generated by our ranking model.} 
\label{fig:contour_favor}
\end{figure}

\begin{figure}
\centering
\subfloat[ImageNet-1K]{

\label{fig:img1k}
\includegraphics[width=\linewidth]{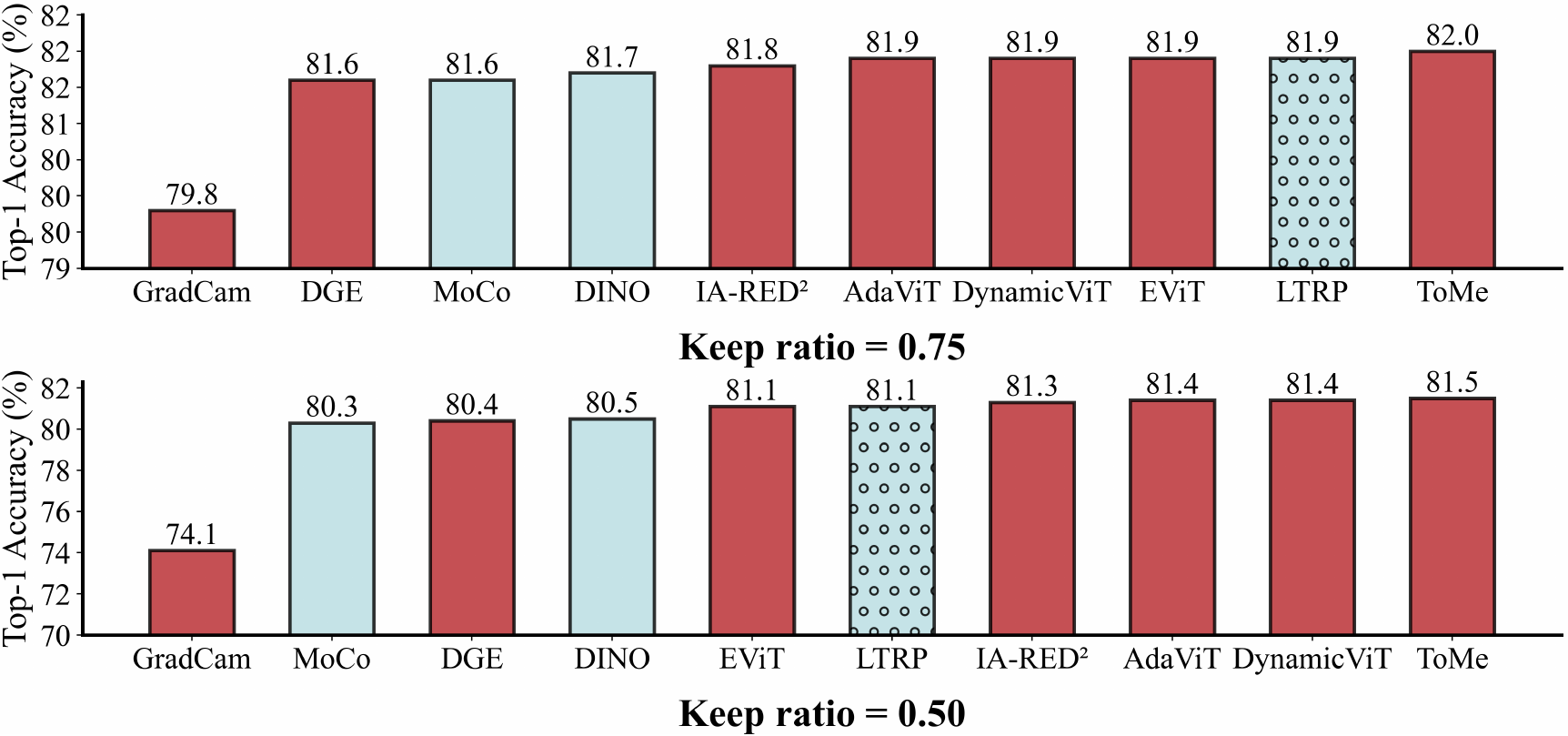}}%
\\

\subfloat[MS-COCO]{
\includegraphics[width=\linewidth]{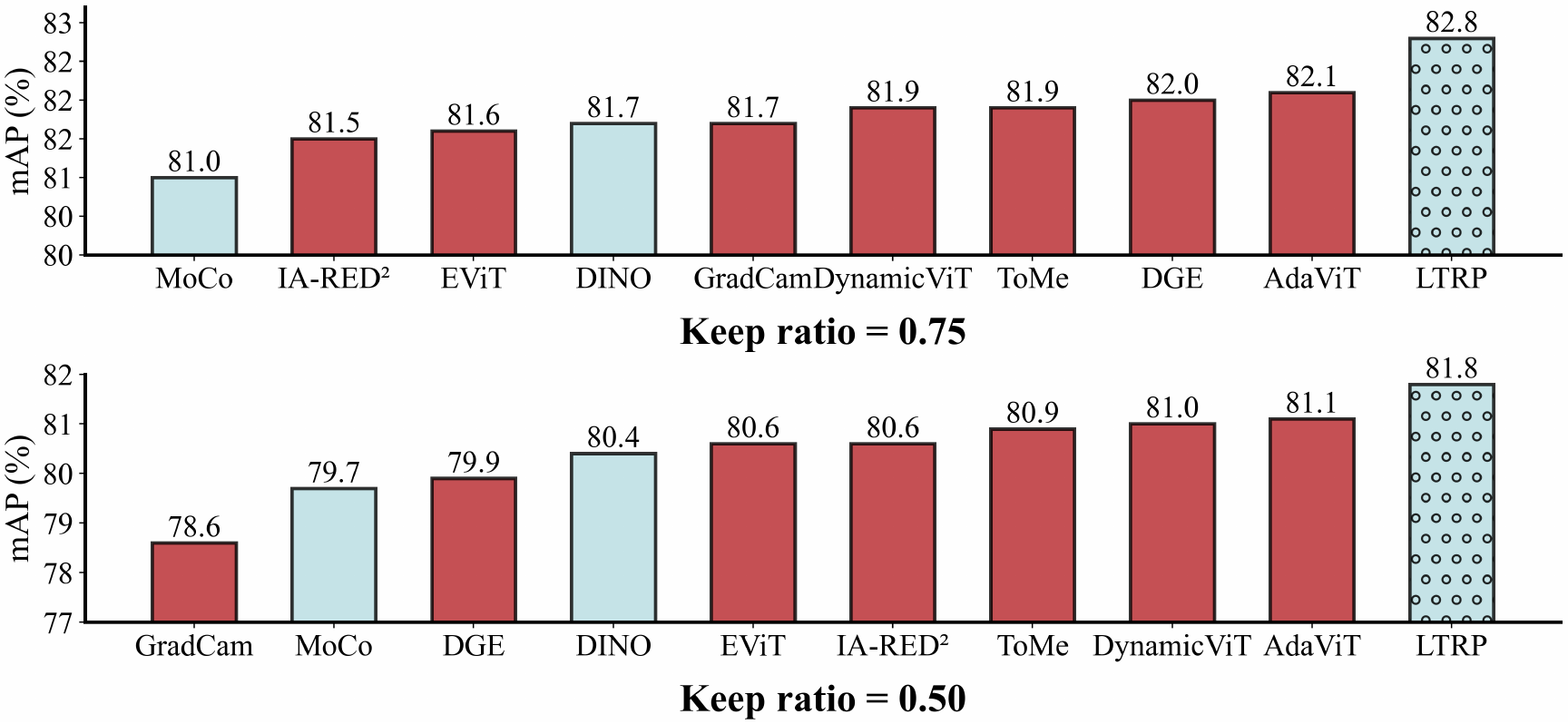}
\label{fig:coco}
}%
\caption{\textbf{Classification results} on single-label dataset (upper, ImageNet-1K) and multi-label dataset (bottom, MS-COCO) at different krs. All methods differ only in the patch selection step. Supervised and self-supervised methods are depicted using different colors. LTRP is annotated spots.}
\end{figure}

\subsection{Learning to Rank}
With pseudo score \(\mathbf{y}\) calculated from \cref{eq2}, we use learning-to-rank algorithms to train an independent ranking model to rank the patches. Formally, each image-partition instance \(\mathbf{x}=<I, \mathbf{P}_v, \mathbf{y}>\) can be viewed as a training sample for the ranking model, which accepts \(\mathbf{x}\) as the input and outputs a vector \(\mathbf{s}=\left[s_1, s_2, ...,s_n\right]\) containing predicted scores for the patches. our LTRP utilizes a list-wise loss, ListMLE \cite{xia2008listwise}, to train the model as follows.
\begin{equation}
L = -log P_s(\pi),  P_s(\pi)=\prod_{i=1}^n \frac{exp\left(s_{\pi(i)}\right)}{\sum_{k=i}^{n} exp\left(s_{\pi(k)}\right)}
\label{eq3}
\end{equation}
where \(\pi\) denotes the descending permutation according to \(\mathbf{y}\), and \(\pi(k)\) is the patch ranked at the \(k\)-th position in \(\pi\). The ranking model will optimize towards sorting patches in line with their semantic density scores. Compared with the other two kinds of ranking loss, i.e., point-wise and pair-wise losses, the list-wise loss more completely considers the overall relationships among patches. In the implementation, we apply a sparse design to the ranking model, which solely inputs the visible patches instead of the full set. We will verify this design and architecture choice of the ranking model in ablation studies.


Note that each training instance only contains a small portion of visible patches and scores. However, when going through a sufficient number of instances, the ranking model sees all patch locations of an image and learns to judge relative ranks among patches. In other words, the ranking model knows how to score all the image patches regarding their informativeness. Consequently, we can solely use the ranking model for image redundancy reduction, as shown in the right part of \cref{fig:idea}.

\subsection{Patch Selection}
\label{sec:patch_selection}

With the ranking model, a straightforward way for redundancy reduction is to select the top-k patches. Our observation suggests that the top patches favor object boundaries rather than homogeneous regions (e.g., dog's body in \cref{fig:intuition}), as shown in \cref{fig:contour_favor}. We attribute it to that changes in object boundaries more easily cause semantic shifts in the reconstructed image. However, a small proportion of patches from homogeneous regions are also essential for visual tasks \cite{liu2023improving}. Thus, we introduce DPC-KNN \cite{du2016study} to capture the representative homogeneous patches. Specifically, to select \textit{k} informative patches including \textit{h} (\textit{h} \(\textless\) \textit{k}) ones from homogeneous regions, we first select (\textit{k} - \textit{h}) patches as described above. Then, we use DPC-KNN to cluster all patches into \textit{k} groups and select the largest \textit{h} ones, given that larger groups are more likely to associate with homogeneous regions. For each selected group, we extract one patch that satisfies the following two conditions, i.e., not selected previously and closest to the cluster center in the remaining patches. The two sets of patches constitute the output of our LTRP. We will ablate \textit{h} in experiments.


\section{Experiments}


\subsection{Datasets and Experimental Setting} 
\textbf{Datasets and evaluation metrics.} We pre-train our LTRP on ImageNet-1K training set, and then assess it on two scenarios as follows. The first is image-level evaluation. We conduct image classification on ImageNet-1K, a single-label classification dataset, and on MS-COCO \cite{lin2015microsoft}, a multi-label classification dataset. The classification model is trained only with the patches remained by different methods. Top-1 accuracy and mAP are employed as the evaluation metrics, respectively. The second is patch-level evaluation. It is conducted both on MS-COCO and PASCAL VOC \cite{Everingham15} with their object detection labels, MS-COCO and ADE20K \cite{zhou2017scene} with their semantic segmentation labels. In practice, we treat all the labeled bounding boxes or masks as foreground and evaluate their consistency with the retained patches by using IoU, F1-score, recall, and precision. Following EViT \cite{liang2022not}, IA-RED\(^2\) \cite{pan2021ia}, etc., we evaluate these methods at different keep ratios (krs). That is, the top-scored \textit{kr}\% patches are preserved. Meanwhile, all the ablation experiments are conducted on ImageNet-100, a subset of ImageNet-1K, given hardware resource constraints.

\textbf{Model architectures.} In LTRP training, we use MAE with ViT-B and choose ViT or CNN (e.g., ResNet50) as our ranking model. In image classification, we utilize ViT-B on ImageNet-1K and MS-COCO, ViT-S on ImgeNet-100 as the backbone for model training. They are both initialized from the MAE encoder \cite{he2022masked}. Moreover, we add ML-Decoder \cite{ridnik2023ml} behind the classification backbone and use Asymmetric loss \cite{ridnik2021asymmetric} for multi-label classification. No additional model is required for the patch-level evaluation, as the performance is obtained by comparing the preserved image patches with groundtruth. All experiments are conducted on a server with 8 NVIDIA GTX 3090 GPUs and Intel(R) Xeon(R) Gold 6226R CPU. 





\begin{figure*}
\centering
\includegraphics[width=1\linewidth]{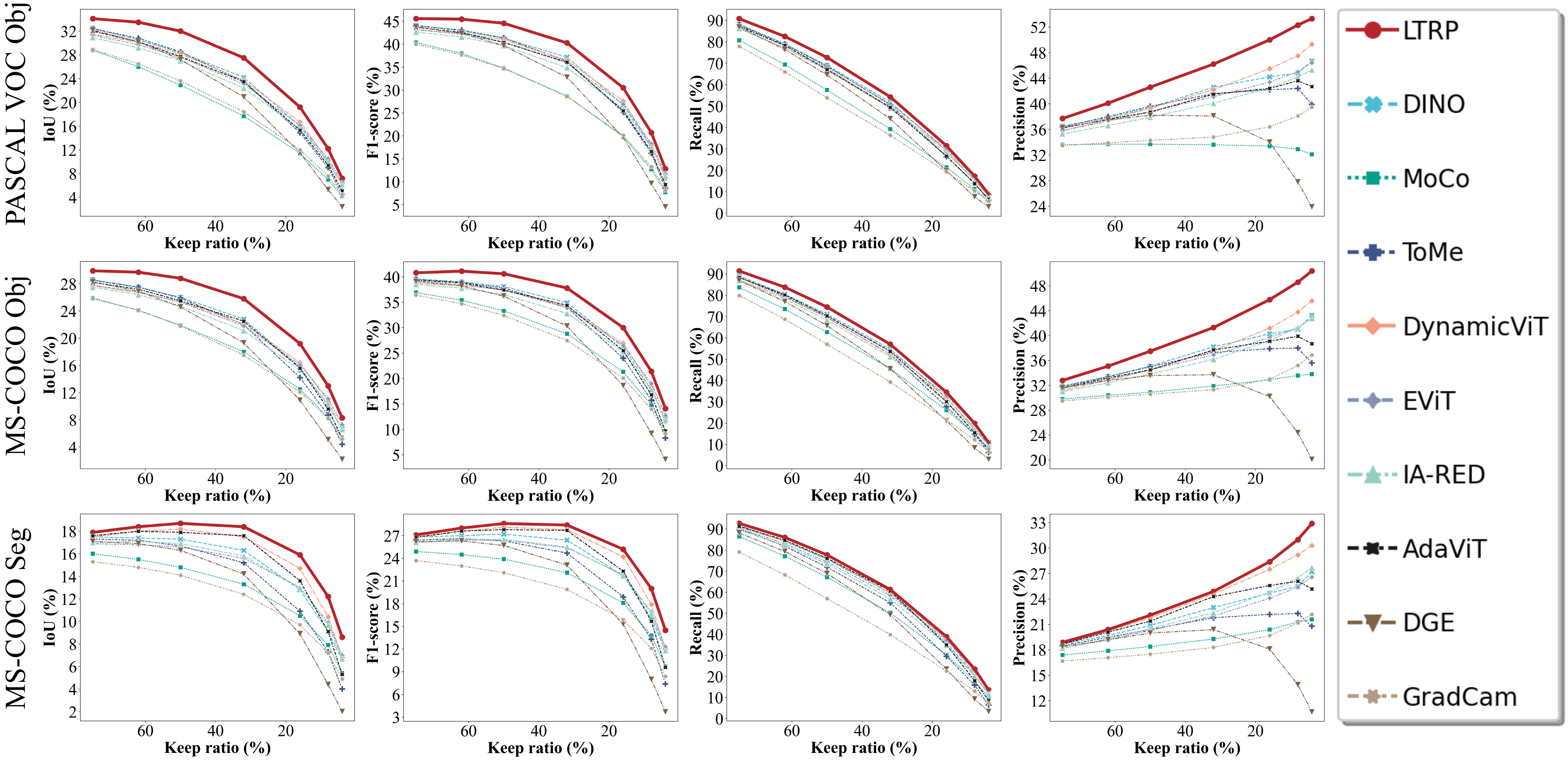}
\caption{
\textbf{Experimental results} on object detection and semantic segmentation datasets. For each dataset, we exclude the categories overlapped with ImageNet-1K (learned) and only compute the metrics on the remaining (unseen) categories, which all the methods have not been explicitly told to learn. The results illustrate the merit of LTRP in unbiased image redundancy reduction.
}
\label{fig:unseen_coco_voc}
\end{figure*}


\textbf{Comparing methods.} 
We compare LTRP with off-the-shelf models from the three classes mentioned previously (see \cref{sec:image_red2}). For CAM, we choose Grad-Cam \cite{selvaraju2017grad} with ViT. For self-supervised methods, we choose MoCo \cite{chen2020improved} with ResNet50 and DINO \cite{caron2021emerging}. For token reduction, we choose EViT \cite{liang2022not}, IA-RED\(^2\) \cite{pan2021ia}, DynamicViT \cite{rao2021dynamicvit}, AdaViT \cite{meng2022adavit}, DGE \cite{song2021dynamic} and ToMe \cite{bolya2022token}. For fair assessment, these methods are all trained (supervised) or pre-trained (unsupervised) on ImageNet-1K. 
 \subsection{Performance Evaluation}


\textbf{Image-level evaluation.} \Cref{fig:img1k} shows the results of different methods at krs 0.7 and 0.5 on single-label classification. The unsupervised and supervised methods are depicted by different colors. LTRP performs better than the self-supervised ones, and slightly worse or on par with state-of-the-art supervised methods. Note that LTRP achieves this without utilizing any supervisory signals. 

We then test LTRP on MS-COCO to validate the performance on multi-label classification. As shown in \cref{fig:coco}, LTRP achieves mAP of 82.8\% and 81.8\% at krs 0.75 and 0.5, respectively. It outperforms the previous best unsupervised methods by 1.1\% and 1.4\%, and the state-of-the-art supervised methods by 0.7\% at both krs. Note that these values correspond to clear accuracy gaps, as the performance differences between existing methods are small. Since different methods only differ in the patches selected, these improvements clearly indicate that LTRP enables a fairer and more meaningful redundancy reduction.

\textbf{Patch-level evaluation.}
To further validate whether the patches are unbiasedly retained, we focus on the categories not included in ImageNet-1K, i.e., the unseen categories, for which all the methods have not been explicitly told to learn. We first evaluate the methods based on the object detection labels. The performance of different methods on PASCAL VOC and MS-COCO is shown in the first two rows in \cref{fig:unseen_coco_voc}. Encouragingly, LTRP surpasses all the competitors in all the evaluated metrics at different krs. The significant improvement in precision highlights that patches selected by LTRP are more accurately located in bounding boxes of unseen categories. With the decrease of krs, the precision of many supervised methods goes up slowly or even goes down. It means that these methods tend to not regard patches from unseen categories as the most important ones, while our LTRP still goes up quickly and successfully avoids this bias. The result implies that LTRP can be well generalized to datasets with quite different categories. 

We then evaluate the methods based on the semantic segmentation labels. The third row in \cref{fig:unseen_coco_voc} shows the results on MS-COCO. LTRP also consistently gives the best results and exhibits a noticeable improvement compared to DynamicViT, the second-best method. We observe that the performance gap becomes larger as the kr decreases. It again demonstrates that LTRP preserves semantics corresponding to the unseen categories well. 


\begin{table*}[htb]
\centering
    \begin{subtable}{.33\textwidth}
        \begin{tabular}{lcc}
model     & kr=0.75                                        & kr=0.5                                          \\ 
\midrule
MAE-B     & {\cellcolor[rgb]{0.89,0.89,0.89}}\textbf{84.4} & {\cellcolor[rgb]{0.89,0.89,0.89}}\textbf{83.3}  \\
MAE-L     & 84.2                                           & 83.2                                            \\
MAE-L-GAN & 84.2                                           & 83.0                                            \\
          &                                                &                                                                                    
\end{tabular}
        \subcaption{\textbf{Reconstruction model.} Lightweight MAE performs better.}
        \label{tab:abalation_mim_model}
    \end{subtable}
    \begin{subtable}{.33\textwidth}
        \centering
        \begin{tabular}{cccc}
\multicolumn{1}{l}{ratio} & speedup & \multicolumn{1}{l}{kr=0.75}           & \multicolumn{1}{l}{kr=0.5}             \\ 
\midrule
0.95                      & \textbf{6.4x}    & 84.3                                  & 83.2                                   \\
0.9                       &  {\cellcolor[rgb]{0.89,0.89,0.89}}3.1x    & {\cellcolor[rgb]{0.89,0.89,0.89}}84.4 & {\cellcolor[rgb]{0.89,0.89,0.89}}83.3  \\
0.8                       & 1.5x    & \textbf{84.8}                         & \textbf{ 83.6}                         \\
0.7                       & 1.0x       & 84.6                                  & 83.4   
\end{tabular}
        \subcaption{\textbf{Masking ratio.} A ratio of 0.9 gets a better trade-off between speedup and accuracy.}
        \label{tab:abalation_mask_ratio}
    \end{subtable}
    \begin{subtable}{.33\textwidth}
        \centering
        \begin{tabular}{lcc}
metrics   & kr=0.75                                        & kr=0.5                                          \\ 
\midrule
\(\ell_1\) & {\cellcolor[rgb]{0.89,0.89,0.89}}\textbf{84.4} & {\cellcolor[rgb]{0.89,0.89,0.89}}\textbf{83.3}  \\
PSNR      & 83.8                                           & 82.5                                            \\
SSIM      & 83.6                                           & 79.9                                            \\
          &                                                &                                                
\end{tabular}
        \subcaption{\textbf{Distance Metric.} The \(\ell_1\) distance is more simple and effective.}
        \label{tab:abalation_distance_metric}
    \end{subtable}%
\\
    \begin{subtable}{.33\textwidth}
        \centering
        \begin{tabular}{lccc}
model    & params                               & kr=0.75                               & kr=0.5                                 \\ 
\midrule
Mobile-  & \multirow{2}{*}{3.2M}                & \multirow{2}{*}{84.3}        & \multirow{2}{*}{83.1}         \\
Former\cite{chen2022mobile}   &                                      &                                       &                                        \\
ViT-T    & 5.5M                                 & \textbf{84.5}                         & \textbf{83.4}                          \\
ViT-S    & {\cellcolor[rgb]{0.89,0.89,0.89}}22M & {\cellcolor[rgb]{0.89,0.89,0.89}}84.4 & {\cellcolor[rgb]{0.89,0.89,0.89}}83.3  \\
ResNet50 & 24M                                  & \textbf{84.5}                         & \textbf{83.4}                         
\end{tabular}
        \subcaption{\textbf{Ranking Model.} All models exhibit similar accuracy. For a fair comparison, we employ ViT-S.}
        \label{tab:abalation_ranking_model}
    \end{subtable}
    \begin{subtable}{.33\textwidth}
        \centering
        \begin{tabular}{ccc}
ratio & kr=0.75                               & kr=0.5                                          \\ 
\midrule
0     & 84.4                                  & 83.3                                            \\
0.1   & \textbf{84.7}                         & 83.4                                            \\
0.2   & {\cellcolor[rgb]{0.89,0.89,0.89}}84.6 & {\cellcolor[rgb]{0.89,0.89,0.89}}\textbf{83.7}  \\
0.3   & 84.6                                  & 83.6                                            \\
1     & 83.6                                  & 82.7                                           
\end{tabular}
        \subcaption{\textbf{Clustering ratio.} A few patches with homogeneous regions can improve classification accuracy.}
        \label{tab:abalation_cluster_ratio}
    \end{subtable}
    \begin{subtable}{.33\textwidth}
        \centering
        \begin{tabular}{lcc}
loss       & \multicolumn{1}{l}{kr=0.75}           & \multicolumn{1}{l}{kr=0.5}             \\ 
\midrule
Regression & 84.1                                  & 83.0                                   \\
RankNet    & 82.9 & 79.5  \\
ListNet    & 84.1                       & 82.5                         \\
ListMLE    & {\cellcolor[rgb]{0.89,0.89,0.89}}\textbf{84.4}                                  & {\cellcolor[rgb]{0.89,0.89,0.89}}\textbf{83.3}       \\
                 &                                                &                                  
\end{tabular}
        \subcaption{\textbf{Ranking loss.} List-wise methods perform better than point-wise and pair-wise ranking losses.
}
         \label{tab:abalation_rank_loss}
    \end{subtable}%
\\
    \begin{subtable}{.33\textwidth}
        \centering
        \begin{tabular}{lccc}
phase     &speedup        & kr=0.75                               & kr=0.5                                 \\ 
\midrule
encoder     & 1   & \textbf{84.4}                         & \textbf{83.7}       \\
decoder         &  {\cellcolor[rgb]{0.89,0.89,0.89}}\textbf{1.3x}   & {\cellcolor[rgb]{0.89,0.89,0.89}}\textbf{84.4} & {\cellcolor[rgb]{0.89,0.89,0.89}}83.3  \\
\end{tabular}
        \subcaption{\textbf{Removal phase.} Removing visible patches before the decoder well balances accuracy and speed.}
        \label{tab:abalation_rmoval_phase}
    \end{subtable} 
    \begin{subtable}{0.33\textwidth}
        \centering
        \begin{tabular}{lcc}
input                                  & kr=0.75                                        & kr=0.5                                          \\ 
\midrule
visible patches & {\cellcolor[rgb]{0.89,0.89,0.89}}\textbf{84.7} & {\cellcolor[rgb]{0.89,0.89,0.89}}\textbf{83.5}  \\
all patches   & 84.4                                           & 83.3                                      
\end{tabular}
        \subcaption{\textbf{Sparse design.} Inputting only visible patches into the ranking model yields better results.}
        \label{tab:abalation_sort_ireelevant}
    \end{subtable}%
    \begin{subtable}{.33\textwidth}
        \centering
        \begin{tabular}{lcc}
case             & kr=0.75                                        & kr=0.5                                          \\ 
\midrule
pixel (w/o norm) & {\cellcolor[rgb]{0.89,0.89,0.89}}\textbf{84.4} & {\cellcolor[rgb]{0.89,0.89,0.89}}\textbf{83.3}  \\
pixel (w norm)   & 81.5                                           & 75.7                                                                                        
\end{tabular}
        \subcaption{\textbf{Reconstruction target.} LTRP favors MIM models based on raw pixels. }
        \label{tab:abalation_reconstruction_target}
    \end{subtable}
    
\caption{\textbf{LTRP ablation experiments} on ImageNet-100. We report accuracy (\%) results for krs \textbf{0.75} and \textbf{0.5}, respectively. The selected configurations are marked in \colorbox[rgb]{0.89, 0.89, 0.89}{gray}.}
\label{tab:ablation}
\end{table*}

\subsection{Ablations}
In \cref{tab:ablation}, we provide detailed ablation experiments and introduce some empirical observations as follows.


\textbf{Reconstruction model.}  \cref{tab:abalation_mim_model} presents LTRP with different MAE as the image reconstruction model. The optimal model is surprisingly the lightweight model MAE-B. Utilizing MAE-B to generate pseudo label boosts the accuracy by 0.2\% compared to larger MAE-L (21 M \textit{v.s.} 81 M in \#params). Moreover, MAE-L with GAN \cite{goodfellow2020generative} loss can get even better reconstruction, but performs worse at kr 0.5. We observe that reconstructions of large models or utilizing GAN loss have finer low-level semantics such as texture details. It tends to pull these patches that excel in recovering low-level semantics ahead in the ranking, consequently reducing the proportion of patches with high-level semantics. This phenomenon causes the accuracy drop.




\textbf{Masking ratio.} ~\cref{tab:abalation_mask_ratio} shows the influence of the ratio of removed patches. The accuracy of an excessively high masking ratio (0.95) drops by 0.1\% compared with baseline. And a lower masking ratio (0.7) does not perform better than the masking ratio of 0.8. We note that extremely high and low masking ratios both result in the issue of \textit{plateaus} in the pseudo score sequence, namely some patches have similar pseudo scores and occupy a continuous position in the ranking sequence. This causes the ranking model to be confused about the right order of these patches. Considering the efficiency issue, LTRP selects a masking ratio of 0.9.

\textbf{Ranking model.}
We investigate the architecture, sparse design and similarity metric of ranking model in \cref{tab:abalation_sort_ireelevant,tab:abalation_ranking_model,tab:abalation_distance_metric}, respectively. We find out that our ranking model is model-agnostic, and even smaller models like ViT-T or ResNet50 can achieve performance similar to the baseline (ViT-S). However, to align with the comparing methods, we select ViT-S. The sparse design excludes the sort-irrelevant patches from the ranking model, it improves accuracy by 0.2\% to 0.3\% across different krs. Moreover, by skipping these patches, we significantly reduce computational costs.

\textbf{Clustering.} \cref{tab:abalation_cluster_ratio} presents our investigation of clustering. Interestingly, selecting all patches (with clustering ratio 1) through clustering at the pixel level is not extremely bad compared with our baseline (82.7\% \textit{v.s.} 83.7\%). A small subset (e.g., 20\%) of patches from homogeneous regions gets an improved LTRP.

\textbf{Ranking loss.} \cref{tab:abalation_rank_loss} studies the choice of ranking loss. Interestingly, the point-wise approach also has decent performance in our LTRP. Compared to ListMLE, its accuracy drops by only 0.3\% at both krs. Point-wise methods train \(\textless\)query, document\(\textgreater\) pair independently, ignoring the relationship among documents. In contrast, our ranking model considers all documents (patches), i.e., \(\textless\)query, document\textbf{s}\(\textgreater\), enabling the acquisition of holistic knowledge. For pair-wise methods, the accuracy drops obviously at krs of 0.75 and 0.5 (82.9 \textit{v.s.} 84.4 and 79.5 \textit{v.s.} 83.3, respectively). Overall, ListMLE performs the best and is also the most appropriate method for utilizing our pseudo-labels.

\textbf{Removal phase.} In \cref{tab:abalation_rmoval_phase}, although the results of removing visible patches before the lightweight decoder show a decrease in performance compared to the approach of removing before the encoder (a reduction of 0.4\% at kr 0.5), we can achieve a noteworthy 1.3x speedup in terms of FLOPs (7.07G \textit{vs.} 5.38G). So we employ the former.

\textbf{Reconstruction target.} Pre-patch normalization is a trick that normalizes a patch using the mean and standard deviation of its pixels. In \cref{tab:abalation_reconstruction_target}, we observe that utilizing pre-trained MAE based on pre-patch normalization for generating pseudo-labels performs exceedingly poorly, which is 7.6\% worse than using unnormalized pixels. We suspect this drop partially because LTRP requires quantifying the global semantic changes between two images. Conversely, pre-patch normalization  appears to reducing global coherence, as it focuses on enhancing the contrast locally \cite{he2022masked}.

\begin{figure}[!htpb]
    \centering
    \includegraphics[width=\linewidth]{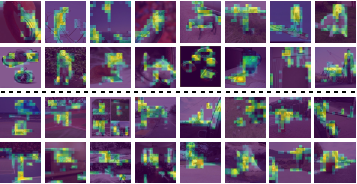}
    \caption{Normalized score maps generated by LTRP on \textit{validation} set of ImageNet-1K (top) and MS-COCO (bottom). Only the top-scored 64 patches are displayed.}
    \label{fig:scores}
\end{figure}

\subsection{LTRP for Efficient ViT}

\begin{table}[!htp]
\centering

 \resizebox{0.49\textwidth}{!}{

\begin{tabular}{lcccccc} 
\toprule
\multirow{2}{*}{cls.}  & \multirow{2}{*}{ranking} & \multicolumn{1}{c}{\multirow{2}{*}{params}} & FLOPs $\downarrow$               & \multicolumn{2}{c}{inference speed$\uparrow$}                      & \multirow{2}{*}{top-1 acc.}    \\
                       &                          & \multicolumn{1}{l}{}                        &                                  & GPU                               & CPU                            &                                \\
model                  & model                    & (M)                                         & (G)                              & im/s                              & im/s                           & \multicolumn{1}{l}{}           \\ 
\hline\hline
\multirow{3}{*}{ViT-B} & -                        & 86                                 & 17.6                             & 1095.1                            & 4.3                            & $83.3^\dagger$)                \\
                       & ViT-S                    & 108\textcolor{red}{(+25.6\%)}               & 13.1\textcolor{blue}{(-25.6\%)}  & 1034.9\textcolor{red}{(-5.5\%)}   & 4.1\textcolor{red}{(-4.7\%)}   & 83.2\textcolor{red}{(-0.1\%)}  \\
                       & ViT-T                    & 91.5\textcolor{red}{(+6.4\%)}               & 13.1\textcolor{blue}{(-25.6\%)}  & 1302.2\textcolor{blue}{(+18.9\%)} & 5.1\textcolor{blue}{(+18.6\%)} & 83.1\textcolor{red}{(-0.2\%)}  \\ 
\hline
\multirow{3}{*}{ViT-L} & -                        & \textcolor[rgb]{0.125,0.129,0.141}{307}     & 61.62                            & 357.7                             & 1.3                            & $85.8 ^\dagger$)               \\
                       & ViT-S                    & 329\textcolor{red}{(+7.2\%)}                & 46.0\textcolor{blue}{(-25.3\%)}  & 445.6\textcolor{blue}{(+24.6\%)}  & 1.5\textcolor{blue}{(+15.4\%)} & 85.7\textcolor{red}{(-0.1\%)}  \\
                       & ViT-T                    & 312.5\textcolor{red}{(+1.8\%)}              & 45.9\textcolor{blue}{(-25.5\%)}  & 485.5\textcolor{blue}{(+35.8\%)}  & 1.7\textcolor{blue}{(+23.1\%)} & 85.6\textcolor{red}{(-0.2\%)}  \\ 
\hline
                       & -                        & \textcolor[rgb]{0.125,0.129,0.141}{632}     & 167.4                            & 140.0                             & 0.5                            & $86.6 ^\dagger$)               \\
ViT-H                  & ViT-S                    & 654\textcolor{red}{(+3.5\%)}                & 124.8\textcolor{blue}{(-25.4\%)} & 181.0\textcolor{blue}{(+29.3\%)}  & 0.6\textcolor{blue}{(+20.0\%)} & 86.3\textcolor{red}{(-0.3\%)}  \\
                       & ViT-T                    & 637.5\textcolor{red}{(+0.9\%)}              & 124.7\textcolor{blue}{(-25.5\%)} & 185.8\textcolor{blue}{(+32.7\%)}  & 0.6\textcolor{blue}{(+20.0\%)} & 86.1\textcolor{red}{(-0.5\%)}  \\
\bottomrule
\end{tabular}
}
\caption{\textbf{Experiments on efficient ViT} on ImageNet-1K at kr 0.75. Different combinations of the ranking and classification (cls.) models show different balances between FLOPs (or inference speed) and accuracy. "-" means do not incorporate the ranking model and select all patches. \(\dagger\) means run by us. When using ViT-T as the ranking model, LTRP accelerates the inference ranging from 18.9\% to 32.7\% with accuracy sacrifice of 0.1\%-0.5\%.}
\label{tab:redundancy_reduction}
\end{table}


LTRP allows the classification model to be trained only on a subset of image patches. We verify whether this LTRP-based classification would be more efficient for ViT.

We elucidate this issue from both inference and training perspectives. For inference, time for both patch ranking and classification should be counted. As shown in \cref{tab:abalation_ranking_model}, even ViT-T has commendable performance. So we choose small ranking model and large classification model as LTRP-based solutions. \Cref{tab:redundancy_reduction} enumerates several combinations on ImageNet-1K at kr 0.75. Using ViT-T as the ranking model shows 1.19\(\times\), 1.36\(\times\), or 1.33\(\times\) acceleration on GPU over the scheme that directly uses ViT-B, ViT-L or ViT-H as the classification model, respectively (ViT-H uses a 16*16 patch grid while ViT-B and ViT-L use 14*14). Meanwhile, the accuracy drops only range from 0.2\% to 0.5\%. The results imply that LTRP-based ones are more appropriate choices in scenarios requiring both efficiency and accuracy. With the popularity of large visual and multi-modal models, large ViT models are more frequently utilized and we believe that these solutions would receive more attention. While for training, LTRP-based solutions include running the ranking model once for patch selection and training the classification model $N$ epochs. The cost raised by the ranking model would be negligible as $N$ increases. Meanwhile, training the classification model would also be accelerated by nearly $1/{kr^2}\times$ as fewer patches are involved.

In sum, compared with the vanilla classification, LTRP-based solutions not only accelerate training but also can improve the inference speed up to over 30\% on GPU with a possible neglectable accuracy degradation. 

\subsection{Visualization}
\begin{figure}
\centering
\includegraphics[width=\linewidth]{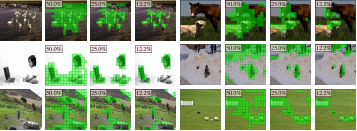}
\caption{Hierarchical redundancy reduction process of LTRP on MS-COCO \textit{validation} set. Keep ratios are shown in the top left of each image. Patches with green boxes are the retained ones.}
\label{fig:different_ratio}
\end{figure}

\Cref{fig:scores} displays the informativeness-aware capability of LTRP on ImageNet-1k and MS-COCO. LTRP well identifies patches located in meaningful object regions in ImageNet-1K images. Meanwhile, it also shows superior unbiasedness in selecting multiple objects from MS-COCO images. We also visualize a process of hierarchical patch selection in \cref{fig:different_ratio}. With the decrease of retained patches, less important patches are gradually removed. LTRP is consistent in favour of patches located on semantic objects. This removal process is mostly in line with human perception. The results again demonstrate the effectiveness of LTRP. 

\begin{figure}

\subfloat[DINO]{
\centering
\includegraphics[width=0.5\linewidth]{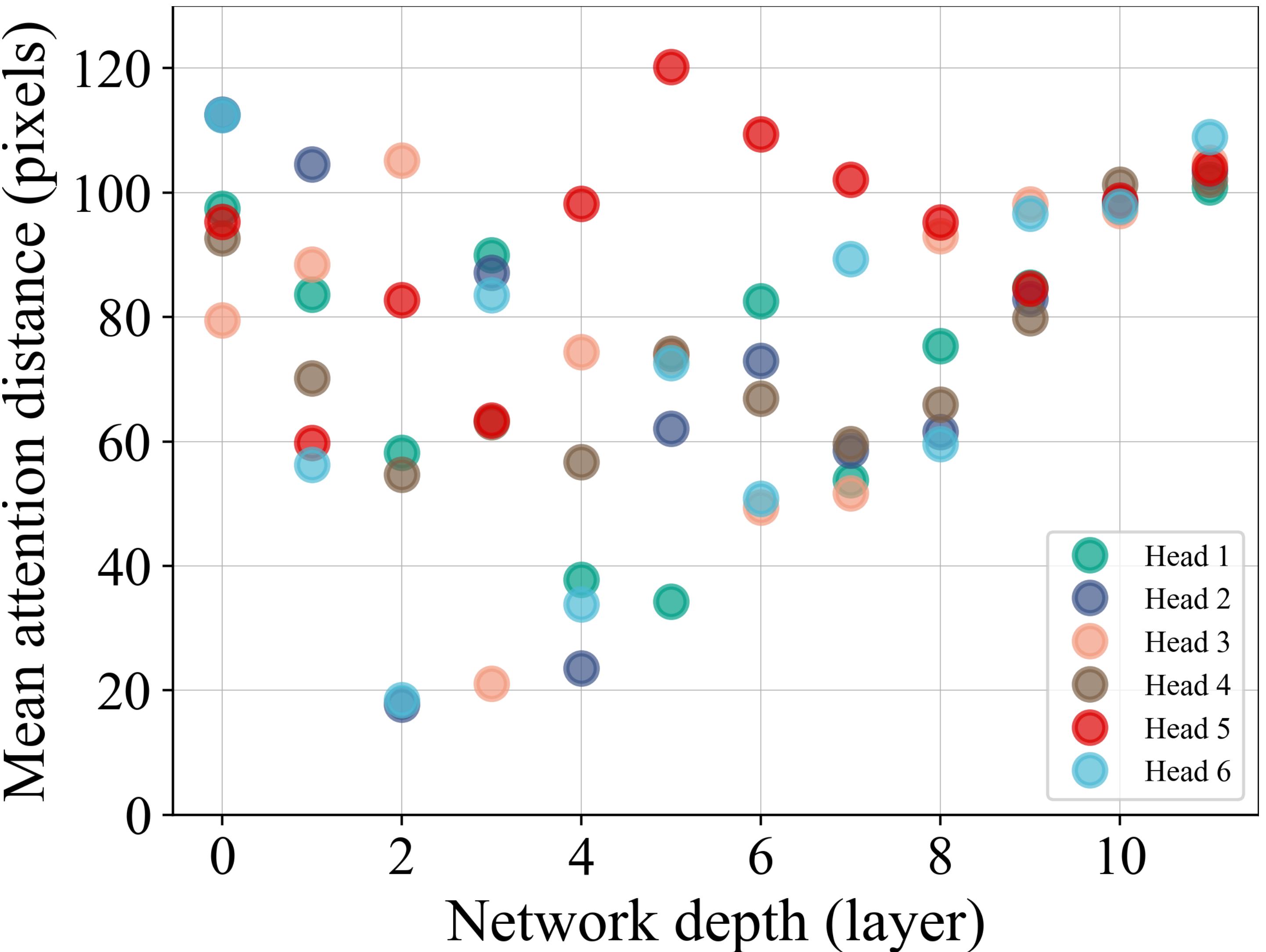}}%
\subfloat[LTRP]{
\centering
\includegraphics[width=0.5\linewidth]{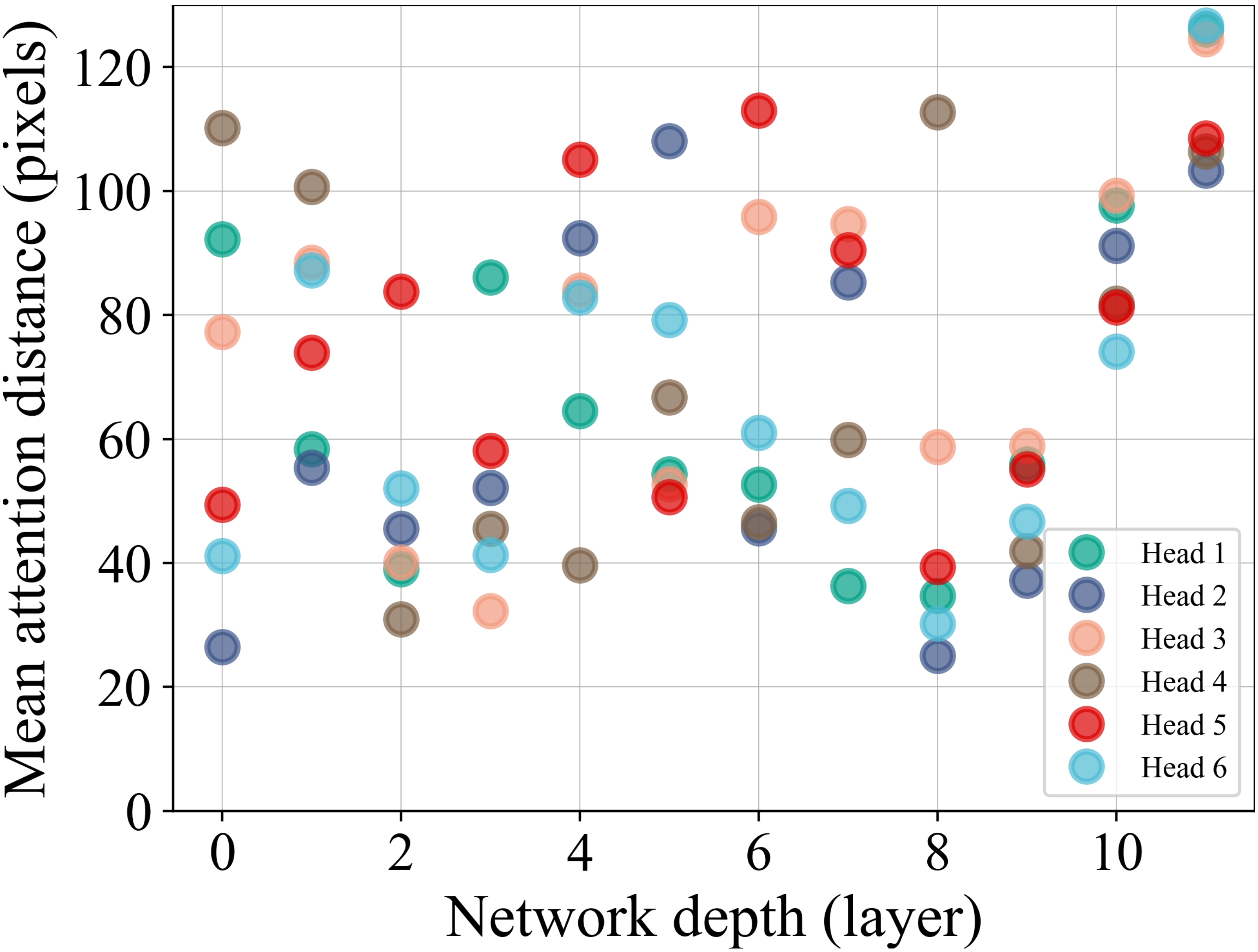}
}%
\caption{ Average attention distances organized according to layer and head for DINO and our LTRP ranking model.}
\label{fig:attention_dis}
\end{figure}

To explore why LTRP performs better than popular self-supervised methods, we plot the average attention distance of each head of DINO and our LTRP ranking model (both are ViT-S) in \cref{fig:attention_dis}. A more scattered distance distribution is observed for LTRP especially in the last few layers. As stated in \cite{wei2022contrastive}, such a distribution suggests that more diverse representation learning is achieved, which explains in part the superiority of LTRP.

\section{Conclusion}
We have introduced LTRP, the first self-supervised method for image redundancy reduction. By leveraging the characteristic of self-supervised learning to create patch-level pseudo labels, and then learning to rank patches, LTRP elegantly eliminates the issue of categorical inductive bias commonly observed in leading supervised methods. Extensive experiments on different datasets and tasks basically validate our proposal. LTRP achieves competitive accuracy in single-label classification and obviously better accuracy in multi-label classification. Meanwhile, the retained patches align more closely with object detection and semantic segmentation labels, especially for categories not explicitly learned by supervised methods. In addition, incorporating LTRP into image classification can lead to a notable speedup in inference with neglectable accuracy degradation if applied appropriately. These results convincingly indicate the effectiveness of LTRP and its unbiased patch selection. We hope that LTRP will provide valuable insights into image redundancy reduction and related visual tasks.

\noindent\textbf{Acknowledgement}
This work was supported by National Science and Technology Major Project (No. 2021ZD0112805) and in part by National Natural Science Foundation of China (No. 62372170, 62172103).
\clearpage
{
    \small
    \bibliographystyle{ieeenat_fullname}      
    \bibliography{main}
}



\end{document}

%% file: preamble.tex
\usepackage[dvipsnames]{xcolor}
